\documentclass[journal]{IEEEtran}
\ifCLASSINFOpdf
\else
\fi

\usepackage{graphicx}
\usepackage{ulem}
\usepackage{amssymb}
 \expandafter\let\csname equation*\endcsname\relax
  \expandafter\let\csname endequation*\endcsname\relax
\usepackage{amsmath}
\usepackage{stmaryrd}
\usepackage{algorithm}
\usepackage{algorithmicx}
\usepackage{program}
\usepackage{color}
\usepackage{epstopdf}

\begin{document}
\title{Canonical Polyadic Decomposition with Auxiliary Information for Brain Computer Interface}

\author{Junhua~Li,~\IEEEmembership{Member,~IEEE,}
        Chao~Li,~
        and~Andrzej~Cichocki,~\IEEEmembership{Fellow,~IEEE}% <-this % stops a space
\thanks{J. Li is with the Laboratory for Advanced Brain Signal Processing, Brain Science Institute, RIKEN, Wako 351-0198, Japan, and also with the Singapore Institute for Neurotechnology (SINAPSE), Centre for Life Science, National University of Singapore, 117456, Singapore (e-mail: {junhua.li@riken.jp; juhalee.bcmi@gmail.com}).}% <-this % stops a space
\thanks{C. Li is with the Department of Information and Communication Engineering, Harbin Engineering University, Harbin 150001, China, and is now working as IPA at the Laboratory for Advanced Brain Signal Processing.}
\thanks{A. Cichocki is with the Laboratory for Advanced Brain Signal Processing,
RIKEN BSI, Wako 351-0198, Japan, and also with the
Systems Research Institute, PAS, Warsaw 00-447, Poland and 
SKOLTECH, Moscow 143026, Russia (e-mail: {a.cichocki@riken.jp}).}}
%\thanks{Manuscript received **; revised **.}}


%\markboth{}
\markboth{IEEE Journal of Biomedical and Health Informatics}%
{J. Li \MakeLowercase{\textit{et al.}}: CPD with Auxiliary Information for BCI}



\maketitle

\begin{abstract}
Physiological signals are often organized in the form of multiple dimensions (e.g., channel, time, task, and 3D voxel), so it is better to preserve original organization structure when processing. Unlike vector-based methods that destroy data structure, Canonical Polyadic Decomposition (CPD) aims to process physiological signals in the form of multi-way array, which considers relationships between dimensions and preserves structure information contained by the physiological signal. Nowadays, CPD is utilized as an unsupervised method for feature extraction in a classification problem. After that, a classifier, such as support vector machine, is required to classify those features. In this manner, classification task is achieved in two isolated steps. We proposed supervised Canonical Polyadic Decomposition by directly incorporating auxiliary label information during decomposition, by which a classification task can be achieved without an extra step of classifier training. The proposed method merges the decomposition and classifier learning together, so it reduces procedure of classification task compared with that of respective decomposition and classification. In order to evaluate the performance of the proposed method, three different kinds of signals, synthetic signal, EEG signal, and MEG signal, were used. The results based on evaluations of synthetic and real signals demonstrated that the proposed method is effective and efficient.  
\end{abstract}


% \fi
\IEEEpeerreviewmaketitle

\section{Introduction}
\IEEEPARstart{B}{rain} Computer Interface (BCI) directly bridges between human brain and the external environment by decoding physiological signals and translating them into understandable information, which has been widely applied to different purposes, such as motor function rehabilitation training  \cite{liu2014tensor} or restoration \cite{pfurtscheller2003thought}, assistive devices \cite{li2013design}, \cite{chai2014brain}, entertainment \cite{li2013competitive}. A key precondition for successful BCI application is the performance of signal decoding (i.e., signal classification) \cite{li2015feature}. Spatial filters, such as independent component analysis and common spatial patterns, have been utilized for feature extraction \cite{zich2015wireless}, \cite{wang2012translation}. In these methods, decomposition is performed based on a matrix, but most of the data existing in nature are presented in a multi-way array (also called tensor). It is better to decompose them based on a multi-way array. Canonical Polyadic Decomposition (CPD) is an important method to decompose the multi-way array into factor matrices \cite{cichocki2009nonnegative}. The factors interact with factors in other dimensions (modes) but not the own dimension. CPD decomposes a tensor with structural information (i.e., preserving interactions between dimensions), unlike the vector-based methods that destroy the relationships among dimensions. CPD has been applied to a number of different kinds of signals including EEG \cite{lee2007nonnegative}, \cite{cichocki2008noninvasive}, \cite{miwakeichi2004decomposing}, \cite{morup2006decomposing}, \cite{de2007canonical}, images \cite{zhao2014bayesian}, videos \cite{zhou2013compact}, fMRI \cite{zhou2013tensor} and so on. All are based on the principle that low-rank factors are used to approximate higher-order tensor (i.e., original tensor used for decomposition) as closely as possible. Those decomposed factors are supposed to preserve intrinsic information while eliminating noise interference. In neurophysiological analyses, factors can be used to detect epileptic location \cite{de2007canonical}, \cite{acar2007multiway} or to reveal underlying mechanisms of brain states \cite{miwakeichi2004decomposing}. In the case of classification, those factors construct an extraction space to obtain features, which are used for the following classifier training. In the above procedure, CPD is utilized in an unsupervised manner to extract the features and neglect label information during decomposition. In this paper, we proposed a supervised CPD method, by which decomposition is better implemented by taking the label information into account. Label information has also been used to seek the optimal subspace based on the criterion of between-class scatter maximization and within-class scatter minimization \cite{liu2014tensor}, \cite{phan2010tensor}, \cite{tao2007general}. The scatter can be measured using different metrics, such as a feature line used in the literature \cite{liu2014tensor}. The data is then projected onto the subspace spanned by the learnt factors, in which the features from different classes have maximal separability. In these literatures, the label information is used at the stage of feature extraction, and the classification is performed by a classifier as the conventional classification way. In our proposed method, the classifier is not required and the feature extraction and classification are merged together.

To demonstrate the performance of the proposed method, we compared it with two state-of-the-arts methods: (1) CPD for feature extraction and support vector machine (SVM) \cite{vapnik2000nature} for classification (2) common spatial patterns (CSP) \cite{soong1995principal}, \cite{ramoser2000optimal} for feature extraction and SVM for classification. The former method is generally used for CPD-based classification \cite{lee2007nonnegative}. The latter method is a classical method used in the brain computer interface, which has a good performance, especially for motor imagery EEG. Many applications with the latter method has been developed up to now \cite{li2013design}, \cite{li2013competitive}. Hence, we compared our method with these two benchmark methods. The datasets used for evaluation consist of synthetic data, real EEG data, and real MEG data. The real EEG and MEG data were acquired from the international BCI competition \footnote[1]{http://www.bbci.de/competition/}.

The rest of paper is organized as follows. Section II describes the proposed method in detail. Section III introduces both synthetic data and real EEG and MEG data used for evaluation. This is followed by comparison results in the Section IV. Then, the discussions are carried out in the Section V. At last, the conclusion is given in the Section VI.
\begin{figure}[!t]
\centering
\includegraphics[width=3.5in]{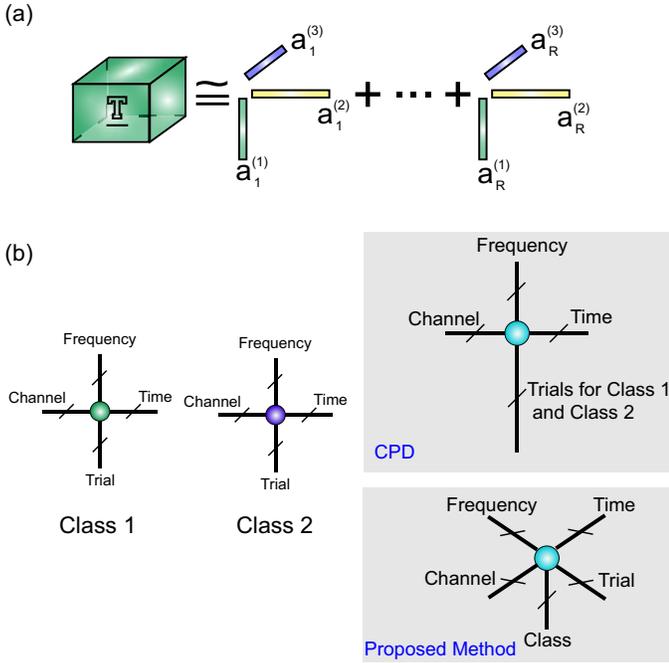}
\caption{(a) CPD for a third-order tensor. (b) Illustration of tensor folding under the condition of two classes for the CPD and the proposed method.}
\label{fig1}
\end{figure}

\section{Method}
\subsection{Notations and Operations}
Vectors are denoted by lowercase letters with boldface, e.g., $\mathbf{a}$. Matrices are denoted by bold capital letters, e.g., $\mathbf{A}$. Higher-order tensors are denoted by bold underlined capital letters, e.g., $\uline{\mathbf{T}}$. The element $(i_1, i_2, \cdots, i_n)$ of a tensor $\uline{\mathbf{T}} \in \mathbb{R}^{I_1 \times I_2 \times \cdots \times I_N}$ is represented by $t_{i_1, i_2, \cdots, i_n}$. $\mathbf{T}_{(n)}$ stands for the mode-n unfolding (matricization) of a tensor $\uline{\mathbf{T}}$. The superscript $\cdot^{T}$ represents transpose, $\succeq$ is symbol for element-wise greater than or equal to, and $vec(\cdot)$ is vectorization operation. The outer product of three vectors $\mathbf{a}^{(1)}\circ \mathbf{a}^{(2)}\circ \mathbf{a}^{(3)}$ forms a rank-one tensor $\uline{\mathbf{T}}$ with entries $t_{i,j,k} = a_i^{(1)}a_j^{(2)}a_k^{(3)}$. The Kronecker product of two matrices $\mathbf{A} \in \mathbb{R}^{I \times J}$ and $\mathbf{B} \in \mathbb{R}^{K \times L}$ is a larger matrix denoted as $\mathbf{A} \otimes \mathbf{B} \in \mathbb{R}^{IK \times JL}$. The Khatri-Rao product of two matrices $\mathbf{A} = [\mathbf{a}_1, \mathbf{a}_2, \cdots, \mathbf{a}_J] \in \mathbb{R}^{I \times J}$ and $\mathbf{B} = [\mathbf{b}_1, \mathbf{b}_2, \cdots, \mathbf{b}_J]  \in \mathbb{R}^{K \times J}$ with the same number of columns $J$ perform the operation $\mathbf{A} \odot \mathbf{B} = [\mathbf{a}_1 \otimes \mathbf{b}_1 ~ \mathbf{a}_2 \otimes \mathbf{b}_2 ~ \cdots ~ \mathbf{a}_J \otimes \mathbf{b}_J]$. CPD decomposes a tensor $\uline{\mathbf{T}}$ into a linear combination of $R$ rank-one terms, denoted by $\uline{\mathbf{T}} \cong \sum\limits_{r = 1}^R {\mathbf{a}_r^{(1)} \circ } \,\mathbf{a}_r^{(2)} \circ \,\mathbf{a}_r^{(3)}$ (Fig. \ref{fig1} (a) illustrates CPD for a third-order tensor). 

\subsection{CPD with Auxiliary Information}
As illustrated in Fig. \ref{fig1} (b), trials for different classes are simply stacked along with the mode of trial in the conventional CPD, but they are kept individual in the proposed method (More details about graphical representation for tensor operations can be found in \cite{cichocki2014era}). An extra dimension is assigned to contain the class. Accordingly, these labels of the classes guide decomposition process during iterations of updating factor matrices. For simplicity and clarity, we explain the proposed method in the case of two classes (see the algorithm pseudocode \ref{alg1}). 

\begin{algorithm}[!b]
\caption{CPD with Auxiliary Information}
\label{alg1}
\textbf{\textit{\textcolor{blue}{Training Stage}}}\\
\textbf{Inputs:} $\uline{\mathbf{T}} $ a fifth-order tensor with size of $I_1 \times I_2 \times I_3 \times I_4 \times I_5$ ($channel \times frequency \times time \times trial \times \ class$)\\
$\mathbf{A}^{(5)}=[1 ~ 0 ; 0 ~ 1]$\\
\textbf{Outputs:} Factor Matrices $\mathbf{A}^{(1)}$, $\mathbf{A}^{(2)}$, $\mathbf{A}^{(3)}$, $\mathbf{A}^{(4)}$\\
\begin{algorithmic} 
\State \textbf{begin}
\State \quad Random initialization for factor matrices 
\State \quad  $\mathbf{A}^{(1)}$, $\mathbf{A}^{(2)}$, $\mathbf{A}^{(3)}$, $\mathbf{A}^{(4)}$
\State \textbf{\qquad repeat}
\State \textbf{\qquad \quad for $n=1$ to $4$ do}
\begin{equation}
\begin{split}
\mathbf{A}^{(n)}=& arg \min \frac{1}{2}\left\| {{\mathbf{A}^{(n)}}{\mathbf{V}^{{{\{ n\} }^T}}} - {\mathbf{T}_{(n)}}} \right\|_F^2\\
&s. ~ t. ~~ \mathbf{A}^{(n)} \succeq 0 \nonumber
\end{split}
\end{equation}
\State \textbf{\qquad \quad end}
\State \textbf{\qquad until} the stopping criterion is met
\State \textbf{end}
\end{algorithmic}

\textbf{\textit{\textcolor{blue}{Testing Stage}}}\\
\textbf{Inputs:} $\uline{\mathbf{D}} $ a fourth-order tensor with size of $I_1 \times I_2 \times I_3 \times 1$ ($channel \times frequency \times time \times trial$)\\
$\mathbf{A}^{(1)}$, $\mathbf{A}^{(2)}$, $\mathbf{A}^{(3)}$\\
\textbf{Outputs:} Classification Label $l$ \\
\begin{algorithmic} 
\State \textbf{begin}
\State \quad 1. Unfolding $\uline{\mathbf{D}} $ into a vector $\mathbf{D}_{(4)} \in  \mathbb{R}^{1 \times I_1I_2I_3}$
\State \quad 2. Unfolding projection space into projection matrix
\State \qquad  $\mathbf{S^{\dag}}=([{\mathbf{A}^{(3)}} \odot {\mathbf{A}^{(2)}} \odot {\mathbf{A}^{(1)}}]^T)^{\dag}$
\State \quad 3. Projection
\State \qquad  $[Pro_1 \quad Pro_2] = \mathbf{D}_{(4)} \mathbf{S^{\dag}}$
\State \quad  4. $l=
\begin{cases}
1& Pro_1 \geq Pro_2\\
2& Pro_1 < Pro_2
\end{cases}$
\State \textbf{end}
\end{algorithmic}

\end{algorithm}

The time series on each channel of each trial are converted into a time-frequency representation by short-time Fourier transform. All time-frequency representations are then assembled to form a fifth-order tensor $\uline{\mathbf{T}} \in \mathbb{R}^{I_1 \times I_2 \times I_3 \times I_4 \times I_5}$ ($channel \times frequency \times time \times trial \times \ class$). The objective of decomposition is to decompose tensor into factor matrices, by which the tensor can be reconstructed approximately. CPD approximates a tensor by $R$ rank-one terms
\begin{equation}
\tilde{\uline{\mathbf{T}}}=\sum\limits_{r = 1}^R {\mathbf{a}_r^{(1)} \circ } \,\mathbf{a}_r^{(2)} \circ \,\mathbf{a}_r^{(3)} \circ \,\mathbf{a}_r^{(4)} \circ \,\mathbf{a}_r^{(5)}. 
\end{equation}
The problem can be modeled to make estimation $\tilde{\uline{\mathbf{T}}}$ as close as possible to original tensor $\uline{\mathbf{T}}$ as follows
\begin{equation}
%\begin{split}
\min \frac{1}{2}\lVert \tilde{\uline{\mathbf{T}}} - \uline{\mathbf{T}} \rVert_F^2 
%&s. ~ t. ~~ \tilde{\uline{\mathbf{T}}} \succeq 0,
%\end{split}
\end{equation}
where $\left\| {\, \cdot \,} \right\|_F$ is Frobenius norm. It means to minimize the square norm of the residual tensor. Furthermore, this minimization problem can be solved by alternately minimizing the mode-n unfolding of the residual tensor. Because the original tensor $\uline{\mathbf{T}}$ consists of time-frequency powers, so it is non-negative. Therefore, we impose non-negativity constraints to factor matrices in order to keep this physical attribute. Then, the minimization can be formulated as
\begin{equation}
\begin{split}
\label{minsubpro}
&\min \frac{1}{2}\left\| {{\mathbf{A}^{(n)}}{\mathbf{V}^{{{\{ n\} }^T}}} - {\mathbf{T}_{(n)}}} \right\|_F^2,\;n = 1, \cdots ,5 \\
&s. ~ t. ~~ \mathbf{A}^{(n)} \succeq 0,
\end{split}
\end{equation}
where,
\begin{equation}
{\mathbf{V}^{{{\{ n\} }^T}}} = {[{\mathbf{A}^{(5)}} \odot  \cdots  \odot {\mathbf{A}^{(n + 1)}} \odot {\mathbf{A}^{(n - 1)}} \odot  \cdots  \odot {\mathbf{A}^{(1)}}]^T}
\end{equation}
$\mathbf{A}^{(n)}$ is factor matrix with non-negative entries corresponding to the mode-n of tensor.

In the proposed method, factor matrix $\mathbf{A}^{(5)}$ is constructed by label information. To this end, we employed orthogonal vectors $[1 ~ 0]^T$ and $[0 ~ 1]^T$, which respectively correspond to class 1 and class 2, to construct the factor matrix $\mathbf{A}^{(5)}$ 
\begin{equation}
{\mathbf{A}^{(5)}} = \left[ {\begin{array}{*{20}{c}}
1&0\\
0&1
\end{array}} \right].
\end{equation}
$\mathbf{A}^{(5)}$  is a constant matrix, so that it is fixed and is not updated while the rest factor matrices are updated at each iteration. For each of the rest four factor matrices, they can be obtained by solving the minimization problem as shown in (\ref{minsubpro}). How to solve the minimization problem (\ref{minsubpro}) is detailed in the \textit{Appendix}.

After all factor matrices $ \mathbf{A}^{(1)},\,\,\mathbf{A}^{(2)},\,\,\mathbf{A}^{(3)},\,\,\mathbf{A}^{(4)},\,\,\mathbf{A}^{(5)} $ are obtained ($\mathbf{A}^{(5)}$ is constant matrix), we can project a new data into subspace spanned by corresponding factor matrices. For example, at the testing stage, there is a new data trial $\uline{\mathbf{D}} \in  \mathbb{R}^{I_1 \times I_2 \times I_3 \times 1}$ (Forth-order tensor $channel \times frequency \times time \times 1$, the last mode corresponds to the trial.). $I_1$, $I_2$, $I_3$, and $1$ are sizes corresponding to channel, frequency, time, and trial, respectively. This trial is first flattened to mode-4 (corresponding to the trial) to form a $1 \times I_1I_2I_3$ vector $\mathbf{D}_{(4)}$ and the projection space is also flattened into a $I_1I_2I_3 \times 2$ projection matrix $\mathbf{S^{\dag}}$ ($\mathbf{S}=[{\mathbf{A}^{(3)}} \odot {\mathbf{A}^{(2)}} \odot {\mathbf{A}^{(1)}}]^T$). Then the classification result of this trial can be obtained by following projection 
\begin{equation}
[Pro_1 \quad Pro_2] = {\mathbf{D}_{(4)}}{\mathbf{S^{\dag}}},
\end{equation}
where $\dag$ stands for the Moore-Penrose pseudo-inverse.
The trial belongs to the class 1 if $Pro_1 \ge Pro_2$, otherwise it belongs to the class 2. Extension to more classes is straightforward. The only change is to set constant matrix $\mathbf{A}^{(5)}$. For example, $\mathbf{A}^{(5)}$ should be set as $[1 ~ 0 ~ 0 ~ 0; ~ 0 ~ 1 ~ 0 ~ 0; ~ 0 ~ 0 ~ 1 ~ 0; ~ 0 ~ 0 ~ 0 ~ 1]$ in the case of four classes. When classifying a trial, classification label is the one corresponding to the largest value.

\section{Evaluation Datasets}
We compared the methods using different kinds of signals: synthetic data and real data. The real data included real EEG signals and real MEG signals, which came from the BCI competition.
\subsection{Synthetic Data}

\begin{figure}[!tb]
\centering
\includegraphics[width=3.5in]{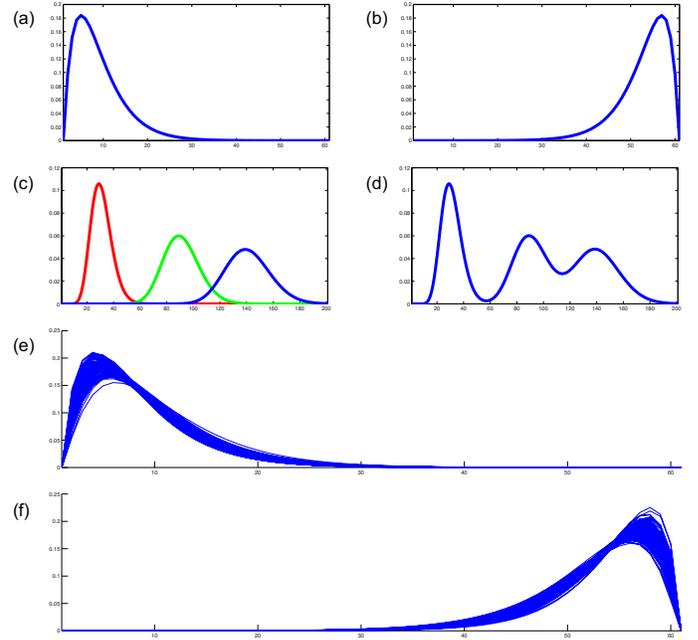}
\caption{(a) The mode-1 course for the class 1. (b) The mode-1 course for the class 2, reversed pattern as that for class 1. (c) Three components for the mode-2. (d) The mode-2 course that is the sum of three components shown in the subplot (c). (e) 100 samples for the mode-1 of class 1. (f) 100 samples for the mode-1 of class 2.}
\label{fig2}
\end{figure}

A third-order tensor was constructed as synthetic data. The mode-1 course for class 1 is shown in Fig. \ref{fig2} (a) while that for class 2 is shown in Fig. \ref{fig2} (b). The mode-2 course is compounded by summing three components (see Fig. \ref{fig2} (c)) up. This results in the course as illustrated in Fig. \ref{fig2} (d). The mode-2 course is the same for both classes. The mode-3 represents trials (samples). Fig. \ref{fig2} (e) shows total 100 samples for the mode-1 of the class 1, and Fig. \ref{fig2} (f) shows samples for the class 2. The mode-1 course is the gamma probability density function, which is determined by two parameters: shape parameter and scale parameter. The values of these two parameters are randomly generated according to a normal distribution with mean of 2 and standard deviation of 0.1. The mode-1 and mode-2 are first formed to be a matrix by $mode$-$1~ \circ  ~mode$-$2$ (61 $\times$ 201), after which Gaussian white noise is added to this matrix. Subsequently, all sample matrices are stacked along the mode-3 to obtain a third-order tensor.  In this case, half of trials (50 trials for each class and the total is 100 trials) were used for training and the rest (100 trials) was for testing.

\begin{figure*}[!bt]
\centering
\includegraphics[width=1\textwidth]{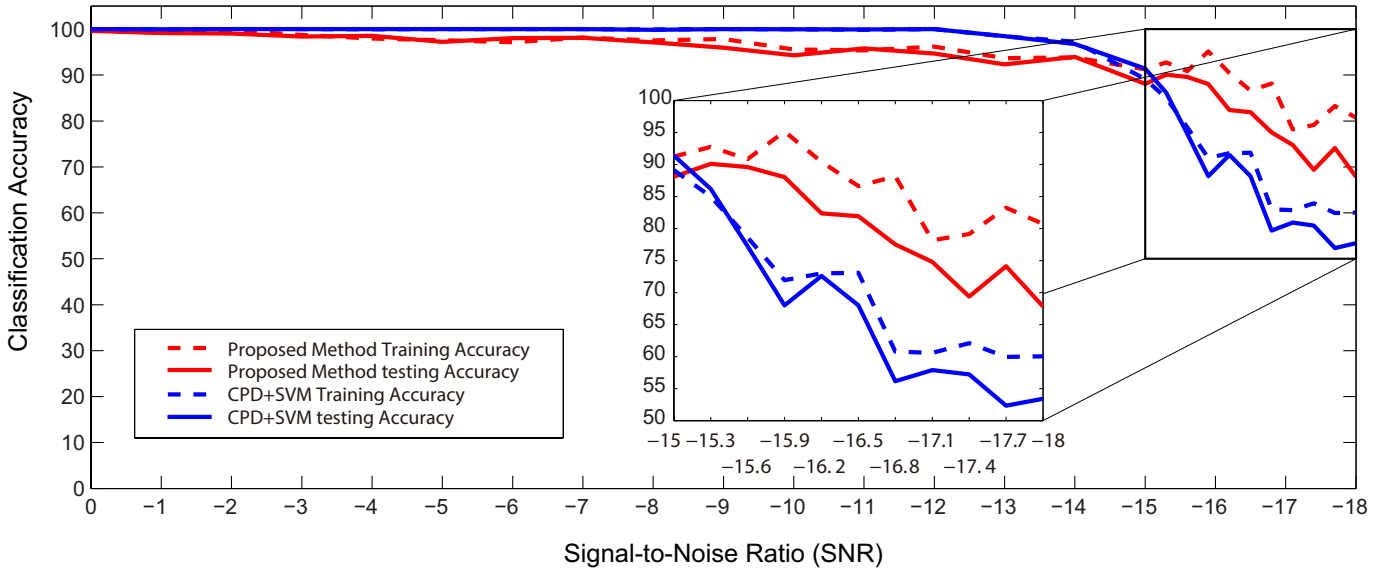}
\caption{Accuracy comparisons between the proposed method and the CPD+SVM method. The horizontal axis shows the signal-to-noise ratios (SNR) while the vertical axis shows the accuracy in percentage. The dotted lines represent training accuracies and the solid lines are for testing accuracies. The red color stands for the proposed method while the blue color stands for the CPD+SVM method.}
\label{fig3}
\end{figure*}

\subsection{Real EEG Data}

The real EEG data are come from international BCI competition II dataset III  (available at https://www.bbci.de/competition/ii/) \cite{schlogl2002estimating}. The experimental protocol was described in \cite{schlogl2002estimating}. The bipolar EEG was recorded with sampling rate of 128 Hz on C3, Cz, and C4 by G.tec amplifier with Ag/AgCl electrodes while a female subject was performing motor imagery with a bar feedback. The task was either left hand motor imagery or right hand motor imagery. The whole recording consists of seven runs, each of which has 40 trials, resulting in 280 trials (the half for training and the other half for testing). The length of a trial is 9 seconds, and the last 6 seconds are within feedback duration. All data have been filtered with a bandpass filter of 0.5$\sim$30 Hz. 

\subsection{Real MEG Data}
The real MEG data came from the S2 of BCI competition IV dataset III (available at http://www.bbci.de/competition/iv/) \cite{waldert2008hand}. MEG signals were recorded at sampling frequency of 625 Hz while healthy subject was moving a joystick from the center position toward one of four potential directions by the right hand and wrist. Trials were partitioned from 0.4 s before to 0.6 s after movement onset and had already been filtered by a band-pass filter (0.5$\sim$100 Hz). Then, data were resampled at 400 Hz. Ten MEG channels on the motor cortex were provided. The number of training trials for each class was 40, resulting in 160 trials totally. The number of testing trials was 73 totally.

\section{Results}
We compared the proposed method with two typical methods on the real EEG and MEG data. One is the conventional procedure when CPD is involved in classification problem, namely CPD for feature extraction and extra classifier for classification. Here, SVM with radial basis function kernel, which is a well-known classifier, was employed. This method is indicated by CPD+SVM in the rest of the paper. The other is the classical method used in the field of brain computer interface, indicating by CSP+SVM in the rest of the paper. We only compared the proposed method with the CPD+SVM on the synthetic dataset, because there were no ERS or ERD phenomena \cite{pfurtscheller2006future} for the synthetic data. The CPD was performed using the codes in Tensorlab \cite{TensorLLaurent}, \cite{sorber2012unconstrained}, \cite{sorber2013structured}. Nonlinear least squares was employed to implement the CPD. The CPD rank was initially set to the number of classes, and then reduced with a step size of 1. The rank with the best performance was finally used. SVM was implemented by MATLAB 2014a build-in commands (MathWorks, U.S.A.) or Libsvm \cite{CC01a}.
%The CPD rank was set to a number between the number of classes and 2 based on their performances.
\subsection{Results for Synthetic Data}

We compared the proposed method with CPD+SVM for different signal-to-noise ratios. The signal-to-noise ratio (SNR) is defined as
\begin{align}
&SNR = \\ \nonumber
&10log_{10}\sqrt{\sum\limits_{{i_1},\,{i_2}, \cdots ,\,{i_N}}^{{I_1},\,{I_2}, \cdots ,\,{I_N}} {t_{{i_1},\,{i_2}, \cdots ,\,{i_N}}^2} /\sum\limits_{{i_1},\,{i_2}, \cdots ,\,{i_N}}^{{I_1},\,{I_2}, \cdots ,\,{I_N}} {n_{{i_1},\,{i_2}, \cdots ,\,{i_N}}^2}},
\end{align}
where ${t_{{i_1},\,{i_2}, \cdots ,\,{i_N}}}$ and ${n_{{i_1},\,{i_2}, \cdots ,\,{i_N}}}$ are the entries of data tensor and noise tensor, respectively.
The accuracy results were plotted in Fig. \ref{fig3} (The stopping criterion for iterations of CPD and the proposed method was established as the difference between two successive iterations or the step size relative to the norm are less than $10^{-12}$). From the Fig. \ref{fig3}, we can see that the performance of the proposed method was slightly lower than that of the CPD+SVM for cases of relatively high SNRs, but it was comparable. When the SNR was low, the proposed method outperformed the CPD+SVM approach. 

\begin{figure*}[!ht]
\centering
\includegraphics[width=0.92\textwidth]{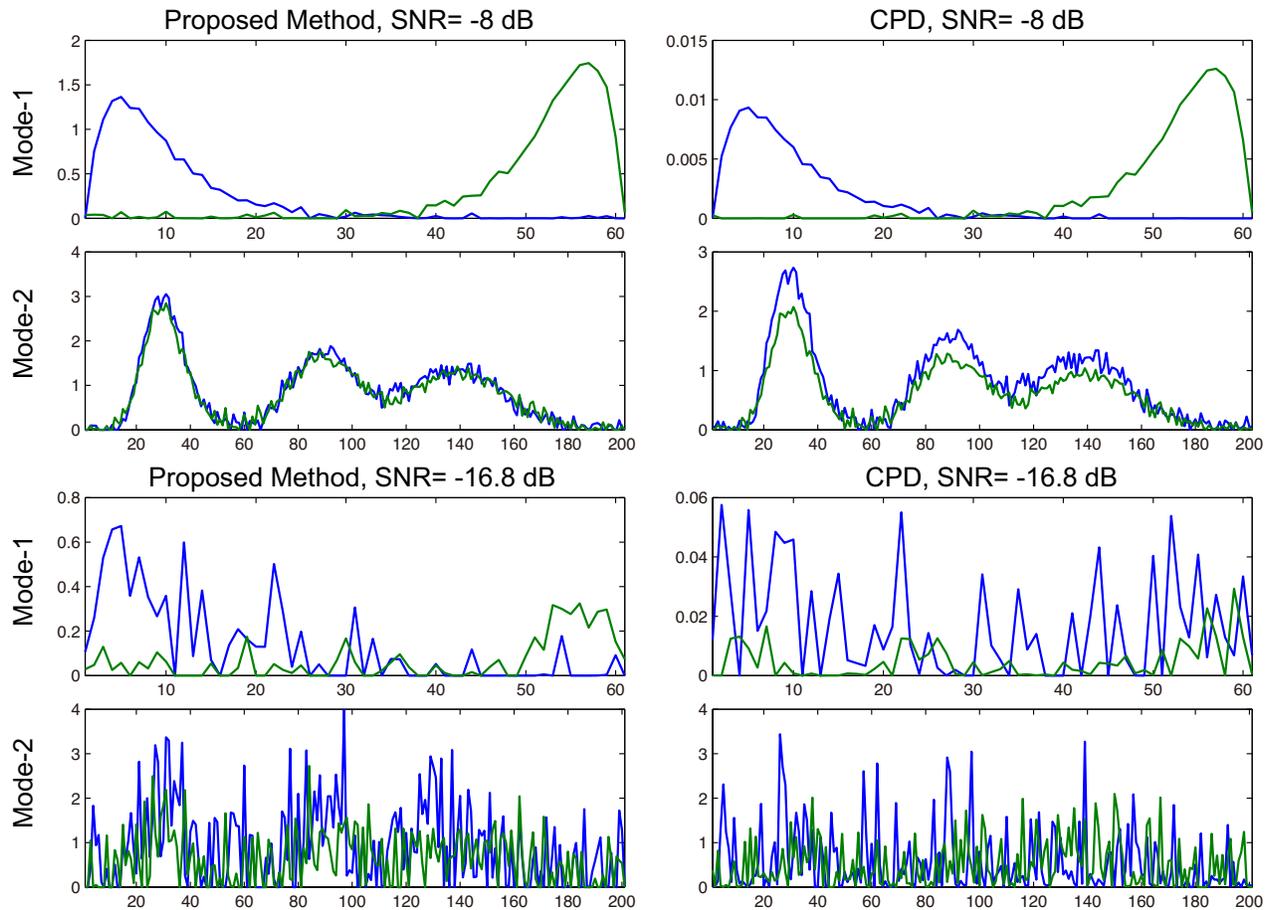}
\caption{The factors corresponding to mode-1 and mode-2 courses. The first column shows factors decomposed by the proposed method, and the second column shows factors decomposed by the CPD for two typical signal-to-noise ratios (SNR= -8 dB that accuracies for both methods are comparable, and SNR= -16.8 dB that the accuracy of the proposed method outperformed that of the CPD+SVM method). Blue lines correspond to the class 1 while green lines correspond to the class 2.}
\label{fig4}
\end{figure*}

\begin{table*}[!ht]
\renewcommand{\arraystretch}{1.3}
\caption{EEG Testing Accuracy Comparisons Between Methods}
\label{table1}
\centering
\scriptsize
\begin{tabular}{|l|c|c|c|c|c|c|c|c|c|c|c|}
\hline
\bfseries Conditions & \bfseries 1 & \bfseries 2 & \bfseries 3 & \bfseries 4 & \bfseries 5 & \bfseries 6 & \bfseries 7 & \bfseries 8 & \bfseries 9 & \bfseries 10 & \bfseries Average (STD) \\
\hline\hline
Proposed Method (Full Band) & 82.14 & 80.71 & 79.29 & 82.14 & 80.71 & 82.14 & 82.14 &81.43 & 80.00 &79.29 & \bf{81.00} (1.17)\\
\hline
CPD+SVM (Full Band) & 46.43    &83.57    &83.57    &83.57    &83.57    &46.43    &46.43    &83.57    &83.57    &46.43 & 68.71 (19.18)\\
\hline
CSP+SVM (Full Band) & - & - & - & - & - & - & - & - & - & - & \bf{80.71}\\
\hline\hline
Proposed Method (8 $\sim$ 21 Hz) & 80.71    &81.43    &80.00    &82.14    &81.43    &81.43    &81.43    &79.29    &80.00    &79.29 & \bf{80.72} (1.01)\\
\hline
CPD+SVM  (8 $\sim$ 21 Hz) &    83.57    &83.57    &83.57    &83.57    &83.57    &83.57    &83.57    &83.57    &83.57    &83.57 & \bf{83.57} (0.00)\\
\hline
CSP+SVM (8 $\sim$ 21 Hz) & - & - & - & - & - & - & - & - & - & - & \bf{80.71}\\
\hline\hline
Proposed Method (1 $\sim$ 7 Hz) &66.43    &65.00    &62.14    &65.00    &64.29    &66.43    &66.43    &66.43    &65.71    &65.71& \bf{65.36} (1.36)\\
\hline
CPD+SVM  (1 $\sim$ 7 Hz) &50.00    &50.71    &50.00    &50.00    &50.00    &50.00    &50.00    &50.71    &50.00  &  50.00 & 50.14 (0.30)\\
\hline
CSP+SVM (1 $\sim$ 7 Hz) & - & - & - & - & - & - & - & - & - & - & 50.71\\
\hline\hline
Proposed Method (22 $\sim$ 30 Hz) & 70.71    &71.43    &71.43    &68.57    &71.43    &68.57    &70.00    &68.57    &69.29    &70.71 & \bf{70.07} (1.24)\\
\hline
CPD+SVM (22 $\sim$ 30 Hz) &56.43    &50.00    &50.00    &56.43    &50.00    &50.00    &50.00    &50.00    &50.00    &50.00& 51.29 (2.71)\\
\hline
CSP+SVM (22 $\sim$ 30 Hz) & - & - & - & - & - & - & - & - & - & - & 55.71\\
\hline
\end{tabular}
\end{table*}

\begin{figure*}[!t]
\centering
\includegraphics[width=0.85\textwidth]{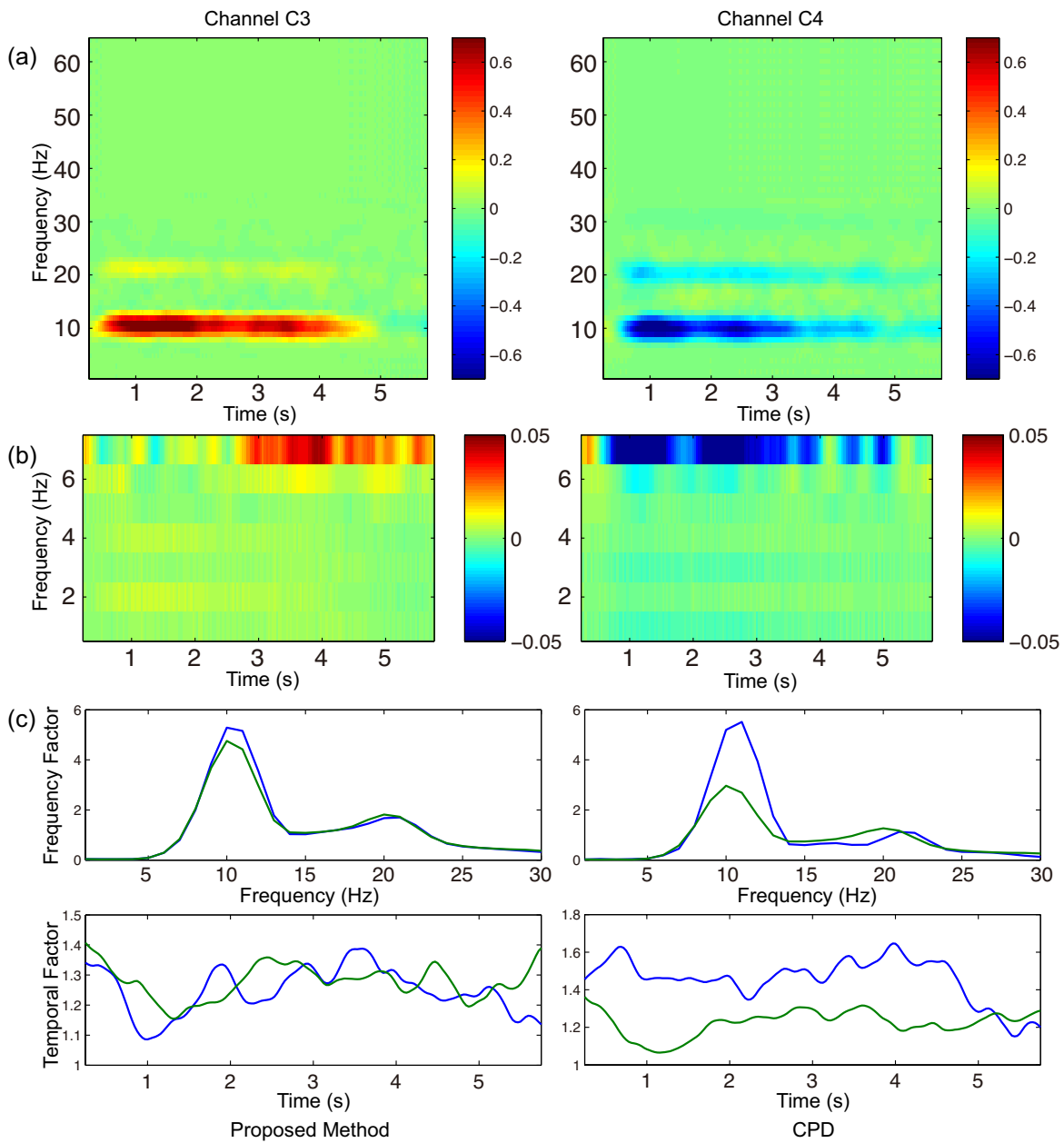}
\caption{The differences of time-frequency representations on channels C3 and C4. The differences are obtained by subtracting the power of right hand motor imagery from the power of left hand motor imagery. (a) Time-frequency difference for full band. (b) Time-frequency difference for band of 1$\sim$7 Hz. (c) Frequency and temporal factors. Factors are depicted in different colors for different classes.}
\label{fig5}
\end{figure*}

In order to look more deeply into the difference between these two methods, we selected two cases of SNRs of -8 dB and -16.8 dB, which are from the comparative performance phase and outperformance phase, respectively. The factors corresponding to mode-1 and mode-2 were depicted in Fig. \ref{fig4}. Compared those factors, we can more intuitively see the reason why the proposed method is better than the CPD+SVM. In the case of SNR -8 dB, both methods could well estimate the real mode-1 and mode-2 courses (refer to Fig. \ref{fig2} (e) and (f) for real mode-1 course, and Fig. \ref{fig2} (d) for real mode-2 course). In the case of the low SNR (SNR= -16.8 dB), the proposed method can estimate them by and large, but the CPD almost fails for estimation. For instance, the mode-1 course estimated for one class by the CPD is always higher than the other class (see subplot of mode-1 for CPD, SNR= -16.8 dB in the Fig. \ref{fig4}), but the truth should be one class higher than the other at the left side and reversed for the other side.

\subsection{Results for Real EEG and MEG Data}

\begin{figure}[!ht]
\centering
\includegraphics[width=3.5in]{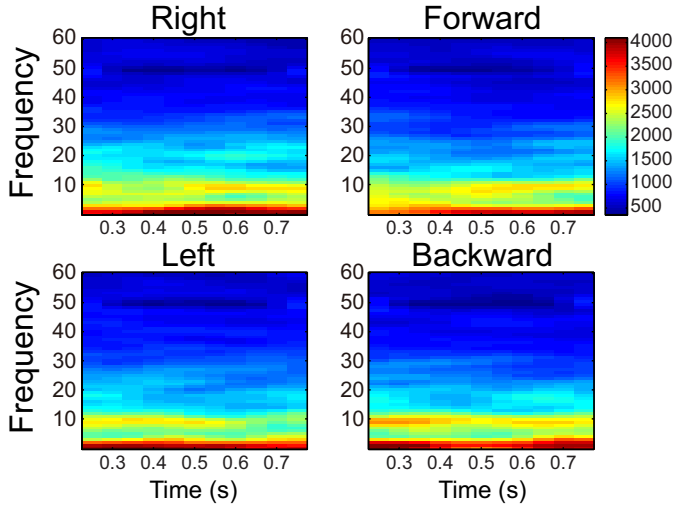}
\caption{The time-frequency representations for four classes (Left, Right, Forward, and Backward) on the channel LC32. Power line frequency (AC) was filtered before drawing the figures.}
\label{fig6}
\end{figure}

\begin{table*}[!ht]
\renewcommand{\arraystretch}{1.3}
\caption{MEG Testing Accuracy Comparisons Between Methods}
\label{table2}
\centering
\scriptsize
\begin{tabular}{|l|c|c|c|c|c|c|c|c|c|c|c|}
\hline
\bfseries Conditions & \bfseries 1 & \bfseries 2 & \bfseries 3 & \bfseries 4 & \bfseries 5 & \bfseries 6 & \bfseries 7 & \bfseries 8 & \bfseries 9 & \bfseries 10 & \bfseries Average (STD)\\
\hline\hline
Proposed Method (All Channels) & 39.73  &  39.73  &  41.10  &  39.73  &  39.73  &  39.73  &  36.99  &  39.73  &  39.73 &   36.99 & \bf{39.32} (1.30)\\
\hline
CPD+SVM (All Channels) & 28.77  &  32.88   & 30.14   & 28.77   & 27.40  &  21.92 &   30.14  &  24.66  &  27.40 &   32.88 & 28.50 (3.41)\\
\hline
CSP+SVM (All Channels) & - & - & - & - & - & - & - & - & - & - & 21.92\\
\hline\hline
Proposed Method (Optimal Channels) & 42.47 &   41.10  &  42.47  &  42.47  &  42.47 &   42.47  &  42.47  &  42.47  &  41.10  &  42.47 & \bf{42.20} (0.58)\\
\hline
CPD+SVM  (Optimal Channels) &    32.88   & 30.14  &  30.14 &   30.14 &   32.88  &  31.51  & 28.77  &  31.51   & 30.14  &  27.40 & 30.55 (1.71)\\
\hline
CSP+SVM (Optimal Channels) & - & - & - & - & - & - & - & - & - & - & 32.88\\
\hline
\end{tabular}
\end{table*}

For the real EEG data, channels C3 and C4 were used, because the hand motor imagery caused bilateral changes on the motor cortex. The part of data with feedback (6 seconds long) for each trial was used in comparison. Fig. \ref{fig5} (a) shows the time-frequency differences between left hand motor imagery and right hand motor imagery on C3 and C4 (The left subtracts the right). The time-frequency distribution clearly shows that the most informative frequency range is between 8 Hz and 21 Hz, within which there are two dominant bands around 10 Hz and 20 Hz, respectively. Beyond that frequency range, the information related to motor imagery is relatively low, but still contains some information (see Fig. \ref{fig5} (b) for 1$\sim$7 Hz in a smaller scale). The stopping criterion for iterations of CPD and the proposed method was that the difference between two successive iterations or the step size relative to the norm are less than $10^{-19}$. The rank used for the CPD is 2. The Fig. \ref{fig5} (c) shows the frequency factors and temporal factors decomposed by CPD and the proposed method, respectively. From the frequency factors, the two dominant bands around 10 Hz and 20 Hz are observed. This matches the fact of EEG characteristics of the dataset.

We compared the results between methods at different frequency bands (i.e., 1$\sim$7 Hz, 8$\sim$21 Hz, 22$\sim$30 Hz, all frequencies). The training trials and testing trials were kept the same as the competition setting. For the decomposed methods, the accuracies varied due to the different random initializations, so we repeated ten times with different initializations for each condition. All testing accuracies are listed in Table \ref{table1}. Numbers enclosed by parentheses are standard deviations. The numbers with boldface are those that have comparable accuracy. In the case of full band (all frequencies), the proposed method achieved the best performance compared with the CPD+SVM and CSP+SVM. The CSP+SVM had a comparable performance as the proposed method done. In the case of the most informative band (i.e., 8$\sim$21 Hz), the best one was the CPD+SVM, but the other two methods were comparable. In the case of the less informative bands (i.e., 1$\sim$7 Hz, and 22$\sim$30 Hz), the proposed method was absolutely preponderant. The CPD+SVM and CSP+SVM methods failed for these frequency bands and their accuracies were close to the chance level (i.e., 50\%).

For the real MEG data, two conditions were compared between methods. One used all ten channels while the other used six channels selected according to the normalized differences between four classes on the training data (this condition is referred to as optimal channels hereafter). As shown in Fig. \ref{fig6}, the dominant frequencies were below 12 Hz, so the band of 0.5 $\sim$ 12 Hz was used for classification. The stopping criterion for iteration of CPD and the proposed method was the same as that used for real EEG data ($<10^{-19}$). The rank used for the CPD is 2. CSP was extended for multiple classes through the manner of one-versus-the-rest. Features were extracted by projecting testing data onto the first largest and smallest eigenvectors. Probability values outputted from SVMs were compared and the label corresponding to the largest probability value was considered as the last classification result. Table \ref{table2} shows the comparison results. The performance of the proposed method was better compared with other two methods not only at the condition of all channels, but also at the condition of the optimal channels.

\section{Discussions}
%\subsection{Robustness of Methods}
Looking at the accuracy results under full band condition for the CPD+SVM, it performed well at some times, but failed at others. This led to low average accuracy. The reason is the CP decomposition largely depends on the initialized values for iterative algorithm. Different initializations could lead to different estimations of factor matrices. Researchers in this field use success rate to count how sensitive the decomposition is to the initialization \cite{acar2011scalable}, \cite{sorber2013optimization}. However, this shortcoming can be overcome by the proposed method in virtue of auxiliary label information during decomposition. Label information constrains decomposition at every iteration so that the difference between classes in the final projection space spanned by the components is maximal. The accuracies only slightly varied among ten runs, and it did not fail for classification at any run. Another point we noticed was that the proposed method performed well not only for the most informative case, but also for the less informative cases. The other two methods (CPD+SVM and CSP+SVM) failed in the cases of the less informative frequency bands. This may be because the proposed method benefits from the supervised decomposition, in which the label information guides the decomposition procedure. The CSP is designed to capture main variance difference between two classes, so it works well under condition of strong difference between classes, but it seems incapable to mine tiny difference of the variance. It is crucial for the decoding method to work well in less informative cases, because spectral band shifting frequently happens and the dominant band might usually be different for different people.

From the results of synthetic data, the accuracies of the CPD+SVM method decreased faster than that of the proposed method with the SNR decreasing. The testing accuracy of the CPD+SVM method was close to the chance level (50\%) in the case of SNR of -16.8 dB, but the proposed method still had an accuracy of near 80\%. It seems that the proposed method is less sensitive to the noise. This point was also demonstrated in the real EEG data for which the proposed method achieved a better performance than others at low informative cases. At these cases, the signal strength related to motor imagery is relatively low.  

The CPD+SVM and CSP+SVM methods implement the classification with two isolated steps: feature extraction step and classification step, whereas the proposed method can achieve classification at single step. The procedure of classification is simplified as the process of independent classifier training is omitted. In addition, the number of parameters of the proposed method is less than that of the CPD+SVM method. Let us consider the case of two classes as an example. Supposing $I_1, I_2, I_3, I_4$ are sizes corresponding to channels, frequencies, time points, and trials of one class, respectively, and $R$ is the rank of the tensor. The number of parameters that CPD needs to estimate is $(I_1+I_2+I_3+2I_4)\times R$, while the number of parameters for the proposed method is $(I_1+I_2+I_3+I_4+2)\times R$. Because the factor matrix corresponding to the mode of class is constant matrix, so the number is reduced to $(I_1+I_2+I_3+I_4)\times R$.

\section{Conclusion}
We proposed a method of physiological signal decoding that feature extraction and classification can be implemented in one step, rather than in two isolated steps. The supervised decomposition is performed with auxiliary label information. In the proposed method, the label information can guide the decomposition to obtain the better factor matrices. The different classes in the space spanned by those factors can be well separated. According to the comparison results for both synthetic data and real data (EEG and MEG), the proposed method possesses better performance. The better performance of signal decoding can promote BCI application and improve the feasibility of BCI. In the future, auxiliary information about other aspects related to neurophysiological responses could be merged to guide the decomposition and may further improve the performance.

% or


\appendices
\section{Solution of minimization problem}
\label{A}
The solution for the minimization problem (\ref{minsubpro}) is detailed below. 
Supposing ${F_{(n)}(\mathbf{a})} \buildrel \Delta \over = {\mathbf{A}^{(n)}}{\mathbf{V}^{{{\{ n\} }^T}}} - {\mathbf{T}_{(n)}}$, where $\mathbf{a}=[\mathbf{a}_1^{(n)^{T}} ~ \mathbf{a}_2^{(n)^{T}} ~ \cdots ~ \mathbf{a}_R^{(n)^{T}}]^T$ is the vectorization of $\mathbf{A}^{(n)}$, the minimization problem (\ref{minsubpro}) can be rewritten as
\begin{equation}
\begin{split}
\label{F1}
&\min \frac{1}{2}\left\| {\,{F_{(n)}(\mathbf{a})}\,} \right\|_F^2,\;n = 1, 2, \cdots ,4 \\
&s. ~ t. ~~ \mathbf{a} \succeq 0.
\end{split}
\end{equation}
We used square mapping $g(\mathbf{a})=[a_1^2 ~ a_2^2 \cdots a_i^2]^T$ to impose the non-negativity constraint, so the minimization problem (\ref{F1}) is in the form of
\begin{equation}
\label{F}
\min \frac{1}{2}\left\| {\,{F_{(n)}(g(\mathbf{a}))}\,} \right\|_F^2,\;n = 1, 2, \cdots ,4.
\end{equation}
$F_{(n)}(g(\mathbf{a}))$ can be represented by Taylor series expansion, which is a sum of infinite terms. In this paper, we used a linear model (namely, first-order model) to approximate $F_{(n)}(g(\mathbf{a}))$ at the location of $\mathbf{a}_k$ in the iteration $k$ as follows
\begin{equation}
\begin{split}
\label{GNM}
&F_{(n)}(g(\mathbf{a})) \approx \\
&vec(F_{(n)}(g(\mathbf{a}_k))) + [\frac{\partial{vec(F_{(n)}(g(\mathbf{a}_k)))}}{\partial{g}} \frac{\partial{g}}{\partial{\mathbf{a}}}]^T(\mathbf{a}-\mathbf{a}_k).
\end{split}
\end{equation}
Substituting (\ref{GNM}) into (\ref{F}) to yield the objective function $m_k(\mathbf{a})$ as
\begin{equation}
\begin{split}
\label{mkf}
&m_k(\mathbf{a}) \buildrel \Delta \over = \\
&\frac{1}{2} \lVert vec(F_{(n)}(g(\mathbf{a}_k))) + [\frac{\partial{vec(F_{(n)}(g(\mathbf{a}_k)))}}{\partial{g}} \frac{\partial{g}}{\partial{\mathbf{a}}}]^T(\mathbf{a}-\mathbf{a}_k) \rVert_F^2.
\end{split}
\end{equation}
The purpose is to update $\mathbf{a}$ making the objective function $m_k(\mathbf{a})$ approach its extremum.
Let $\mathbf{f}_k=vec(F_{(n)}(g(\mathbf{a}_k)))$, $\mathbf{J}_k=[\frac{\partial{vec(F_{(n)}(g(\mathbf{a}_k)))}}{\partial{g}} \frac{\partial{g}}{\partial{\mathbf{a}}}]^T$, and $\mathbf{p}_k=(\mathbf{a}-\mathbf{a}_k)$, substituting them into the objective function $m_k(\mathbf{a})$ 
\begin{eqnarray*}
m_k(\mathbf{p}_k+\mathbf{a}_k) &=& \frac{1}{2}\left\| {{\mathbf{f}_k} + {\mathbf{J}_k}\mathbf{p}_k} \right\|_F^2\\
&=& \frac{1}{2}tr[{({\mathbf{f}_k} + {\mathbf{J}_k}\mathbf{p}_k)^T}({\mathbf{f}_k} + {\mathbf{J}_k}\mathbf{p}_k)]\\
&=& \frac{1}{2}tr[{\mathbf{f}_k}^T{\mathbf{f}_k} + {\mathbf{f}_k}^T{\mathbf{J}_k}\mathbf{p}_k + {\mathbf{p}_k^T}\mathbf{J}_k^T{\mathbf{f}_k} + {\mathbf{p}_k^T}\mathbf{J}_k^T{\mathbf{J}_k}\mathbf{p}_k]\\
&=& \frac{1}{2}\left\| {{\mathbf{f}_k}} \right\|_F^2 + {\mathbf{p}_k^T}\mathbf{J}_k^T{\mathbf{f}_k} + \frac{1}{2} {\mathbf{p}_k^T}\mathbf{J}_k^T{\mathbf{J}_k}\mathbf{p}_k.
\end{eqnarray*}
%\begin{equation}
%\end{equation}
Setting partial derivative of $m_k(\mathbf{p}_k+\mathbf{a}_k)$ with respect to $\mathbf{p}_k$ as zero
\begin{eqnarray*}
\frac{{\partial \,m_k(\mathbf{p}_k+\mathbf{a}_k)}}{{\partial \,\mathbf{p}_k}} = \mathbf{J}_k^T{\mathbf{f}_k} + \mathbf{J}_k^T{\mathbf{J}_k}\mathbf{p}_k &=& 0 \\
\mathbf{J}_k^T{\mathbf{J}_k}\mathbf{p_k} &=&  - \mathbf{J}_k^T{\mathbf{f}_k} \\
\mathbf{p}_k &=&  - {(\mathbf{J}_k^T{\mathbf{J}_k})^{ - 1}}\mathbf{J}_k^T{\mathbf{f}_k}.
\end{eqnarray*}
%\begin{equation}
%\end{equation}
During the updating, a trust-region strategy is utilized to determinate whether $\mathbf{a}$ is updated \cite{conn2000trust}. The $\mathbf{a}$ is updated if the ratio (defined as trustworthiness) of the actual reduction and the predicted reduction is trusted enough. The radius of trust region is updated at every iteration. 
When updated, the mode-n factor matrix can be updated by 
\begin{equation}
\mathbf{a}_{k+1}=\mathbf{a}_k+\mathbf{p}_k  %\nonumber
\end{equation}

\ifCLASSOPTIONcaptionsoff
  \newpage
\fi



%\bibliographystyle{IEEEtran}
%\begin{thebibliography}{1}
%\end{thebibliography}

\bibliographystyle{IEEEtran}
\bibliography{mybibfile}

\end{document}